\documentclass[conference]{IEEEtran}
\IEEEoverridecommandlockouts
\usepackage{cite}
\usepackage{amsmath,amssymb,amsfonts}
\usepackage{algorithmic}
\usepackage{graphicx}
\usepackage{textcomp}
\usepackage{xcolor}
\usepackage{multirow}
\usepackage{booktabs}
\def\BibTeX{{\rm B\kern-.05em{\sc i\kern-.025em b}\kern-.08em
    T\kern-.1667em\lower.7ex\hbox{E}\kern-.125emX}}
\begin{document}

\title{ChatPCG: Large Language Model-Driven Reward Design for Procedural Content Generation \\
}

\author{\IEEEauthorblockN{In-Chang Baek\textsuperscript{1}, Tae-Hwa Park\textsuperscript{2}, Jin-Ha Noh\textsuperscript{2}, Cheong-Mok Bae\textsuperscript{2}, Kyung-Joong Kim\textsuperscript{1,2,*\footnote{corresponding author}}}
\IEEEauthorblockA{
\textit{\textsuperscript{1}AI Graduate School \textsuperscript{2}School of Integrated Technology} \\
\textit{Gwangju Institute of Science and Technology (GIST), South Korea} \\
\{\texttt{inchang.baek@gm., taehwa-p@gm., noah9905@gm., cmbae0307@gm., kjkim@}\}{gist.ac.kr} \\
}
}

% \IEEEoverridecommandlockouts
% \IEEEpubid{\makebox[\columnwidth]{979-8-3503-5067-8/24/\$31.00~\copyright2024 IEEE \hfill} %–> insert the copyright option applicable from above.
% \hspace{\columnsep}\makebox[\columnwidth]{ }}
% % Add the following code after the \\make title command:
% \IEEEpubidadjcol

\maketitle

\begin{abstract}
Driven by the rapid growth of machine learning, recent advances in game artificial intelligence (AI) have significantly impacted productivity across various gaming genres. 
Reward design plays a pivotal role in training game AI models, wherein researchers implement concepts of specific reward functions.
However, despite the presence of AI, the reward design process predominantly remains in the domain of human experts, as it is heavily reliant on their creativity and engineering skills.
Therefore, this paper proposes ChatPCG, a large language model (LLM)-driven reward design framework.
It leverages human-level insights, coupled with game expertise, to generate rewards tailored to specific game features automatically.
Moreover, ChatPCG is integrated with deep reinforcement learning, demonstrating its potential for multiplayer game content generation tasks.
The results suggest that the proposed LLM exhibits the capability to comprehend game mechanics and content generation tasks, enabling tailored content generation for a specified game.
This study not only highlights the potential for improving accessibility in content generation but also aims to streamline the game AI development process.
\end{abstract}


\section{Introduction}
Advances in game artificial intelligence (AI) have witnessed rapid growth within the gaming industry, profoundly impacting productivity across various gaming genres.
Furthermore, machine learning algorithms have emerged as the primary method for training agents in gameplay and for game content generation.
The reward function design plays a crucial role in effectively training AI agents using search-based methods \cite{stephenson2019agent} and deep reinforcement learning (DRL) techniques \cite{khalifa2020pcgrl}.
However, the process of reward function design often relies heavily on researchers' insights and involves iterative trial-and-error approaches.
The reliance of the design process on expert knowledge not only erects barriers to access but also poses challenges to the attempts at designing reward function.

Recently, several approaches have been explored to improve decision making models with large language models (LLMs).
In the realm of robotics, novel approaches have been proposed to leverage domain-specific knowledge and coding capabilities of LLMs for solving the reward generation problem \cite{ma2023eureka}.
These endeavors not only diminish the reliance of the reward design on experts but also enhance the performance of DRL models.
In game content generation literature, there is a growing focus on generating LLM-based content, primarily emphasizing the utilization of LLM as either a data-driven model \cite{todd2023level,sudhakaran2024mariogpt} or a standalone generation model \cite{taveek2023chatgpt4pcg, todd2024missed}.
Meanwhile existing studies proposed  lots of data-free generative algorithms such as search- and learning-based, which do not require training datasets and use objective functions instead; however, no attempts have been made to utilize LLMs for this approach.
Therefore, exploring the incorporation of LLMs with data-free generative algorithms to maximize their utility is necessary.

To address this challenge, this study proposes a LLM-driven reward design framework, \textit{ChatPCG}.
This framework automatically identifies design insights and generates reward functions for specific game environments, leveraging the expertise of LLMs on game mechanism.
Additionally, the proposed method is incorporated into a DRL-based content generation method \cite{khalifa2020pcgrl} to train an agent using the LLM-designed reward function.
To enhance the quality of the reward function, which encompasses complex game elements, we modularize the function into idea units and subdivide the feedback.
The concept-level code is adapted for a specific game through iterative self-alignment processes using game log data.
Thus, this framework is designed to enhance transparency in the reward generation process and facilitate maintenance, thereby making it suitable for the game AI development process.
The effectiveness of ChatPCG is investigated by improving rewards for a multiplayer content generation task \cite{jeon2023raidenv}.
The results show that LLMs can provide human-level insights by considering game characteristics, and the generated content also reflects these design insights.

\section{Background}

% \subsection{Reward Functions in PCGML}

\subsection{Reward Generation Task}
LLMs have been rapidly developed, along with ChatGPT. Particularly, in reinforcement learning, these models have been utilized for task description and code generation. However, since LLMs are insufficiently precise for generating reinforcement learning actions, they have been widely used to create reward functions. The authors of a previous study \cite{yu2023language} showed that by inputting detailed environment descriptions and rules for robotic tasks, LLMs can generate a reinforcement-learning reward function. Meanwhile, compared with humans, \textit{Eureka} \cite{ma2023eureka} demonstrated the ability to generate superior reward functions for complex robotic tasks via evolutionary search and reward reflection. 

Research on prompt engineering (PE) techniques has improved the inference capabilities of LLMs by designing entity of thought.
Traditionally, the Input-Output (IO) method was only suitable for simple tasks that do not require logical reasoning.
Therefore, to address these limitations, the Chain of Thought (CoT) \cite{wei2022chain} has enhanced the reasoning capabilities of LLMs.
CoT was especially employed to modify the reward function sequentially to generate a more sophisticated output.
Meanwhile, another study \cite{zeng2023learning} obtained trajectories from the policy and replay buffer. Iterative self-alignment was then conducted to update the reward function.
Therefore, in this study, reward function components are modularized, and specific feedbacks for alignment are utilized, thereby increasing both the framework explainability and maintainability.

%\subsection{RaidEnv I: Homogeneous Character Generation}
% \input{background/RaidEnv1.tex}

\section{Preliminaries}
\subsection{RaidEnv II: Heterogeneous Character Generation}
\begin{figure}[!h]
    \centering
    \includegraphics[width=0.9\linewidth]{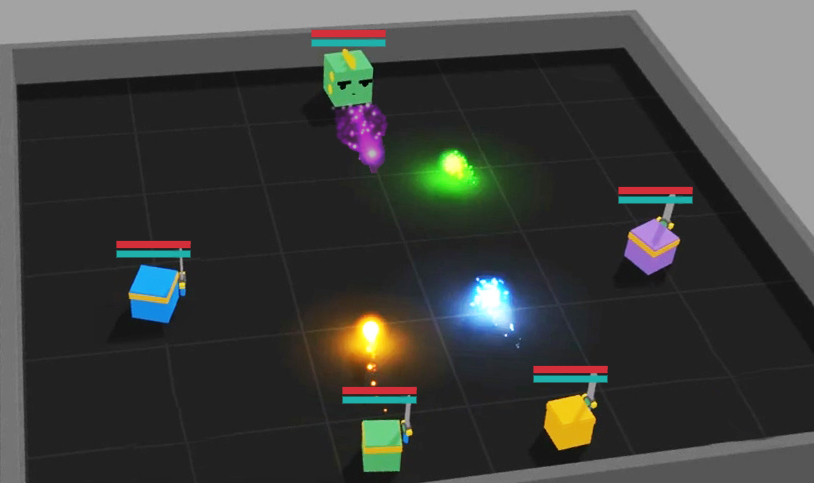}
    \caption{Snapshot of the RaidEnv II environment. The four agents on the bottom-side are players and top-side green agent is the boss.}
    \label{fig:raidenv_snapshot}
\end{figure}

RaidEnv \cite{jeon2023raidenv} is a multiplayer game environment simulating boss raid scenarios, where multiple player agents collaborate to defeat a single boss agent within a limited playtime.
RaidEnv II (Fig. \ref{fig:raidenv_snapshot}) is an updated version that accommodates heterogeneous agent settings.
The player and boss agents possess unique characteristics, skills, and attributes, enhancing the diversity and realism of raid scenarios.
This enhancement fosters various gameplay experiences, more closely mimicking actual raid situations.
This work utilizes seven representative game properties: maximum health, armor, speed, cooldown, cast time, range, and damage values.

RaidEnv II employs a DRL-based content generation framework (i.e., PCGRL \cite{khalifa2020pcgrl}) to generate heterogeneous character configurations.
The \textbf{state} space ($S$) representation consists of four-stacked 44 scalar values, 11 values encompass normalized skill values, skill type, and an update flag for each agent (11 values $\times$ 4 agents). 
The configuration change occurs in a round-robin manner among agents for each step, with the agent being currently updated indicated by the update flag.
The \textbf{action} space ($A$) comprises seven discrete actions which has five action categories (i.e., $O(A)=5^{7}$), allowing for property adjustments to revise agent characteristics and abilities.
While a previous work \cite{jeon2023raidenv} employed a reward function \textbf{reward} focused on controllable generation to achieve desired winrates, multiplayer game aspects were not considered. This study addresses this gap by proposing a RaidEnv II tailored reward function to leverage the ChatPCG framework.

\section{Proposed Method}
\label{sec:method}

\subsection{ChatPCG Framework}
\begin{figure*}[th]
    \centering
    \includegraphics[width=1.0\linewidth]{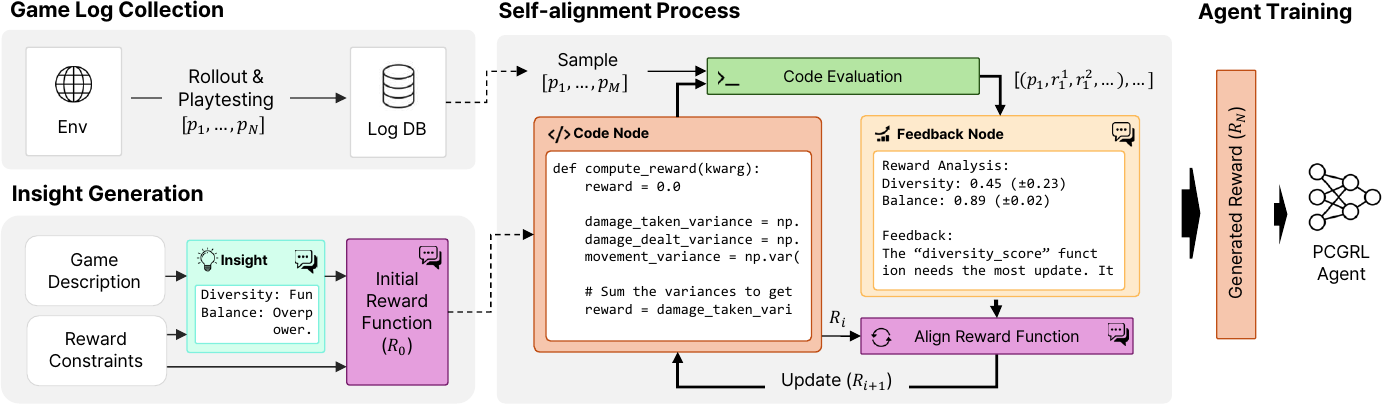}
    \caption{Architecture of ChatPCG framework. “Message icons” indicate the use of LMs in the context. Refer to Section \ref{sec:method} for detailed description.}
    \label{fig:architecture}
    \vspace{-0.34cm}
\end{figure*}

ChatPCG reward generation follows a two-step sequence approach: (1) conceptualize insights into design rewards, and (2) implement these insights into code, then align them with the environment. 
This framework, which has multiple steps, is designed to improve the quality of responses generated by language models (LMs).
Fig. \ref{fig:architecture} illustrates the comprehensive architecture of the proposed framework.

\subsubsection{Insight to Reward Generation}

The process entails steps to establish an initial reward function based on conceptual insights.
The insight generation process receives the textual description of the game environment and reward constraints as input, including the available variables and desired output range of the function.
LMs generate multiple insights to enhance the quality of game contents within the available variables.
These insights are specifically organized into several modules to enhance clarity and maintenance capabilities of reward function during the alignment procedure. 
Subsequently, LMs generate the initial reward function, denoted as $R_{0}$, by implementing design insights into module-level functions. $R_{i}$ represents the reward function at the $i$-th iteration.

\subsubsection{Self-alignment Process}
This solidifies the reward function draft tailored to a specific game environment which is facilitated by employing game log data.
In code evaluation, the module-level and summed reward values are returned by inputting the $M$ rows of game playtested log data, $p_{0:M}$, into the current reward function.
The set of calculated reward values is denoted by $r_{0:M}^{j}$, where $j$ represents each reward function module.
Subsequently, LMs analyze the descriptive results of reward values and generate the textual feedback to revise the reward function.
By evaluating the output of the function using real game values, LMs verify that design insights are effectively reflected in the code.
LMs subsequently generate the next reward function, $R_{i+1}$, along with the served textual feedback and current reward function, $R_{i}$.
This process is repeated $N_{align}$ times and ensures that the reward function is procedurally modified to the weight or formula of the equation to utilize variables in the game accurately.

The proposed self-alignment process is structured based on the CoT method, aiming to alleviate the hallucination of LLMs.
The structuring of the code revision process in a chain allows LLMs to generate a single feedback at each step, pinpointing an improvement point accurately.
The efficiency of the self-alignment process was investigated by comparing the zero-shot code generation (i.e., IO) with the CoT-based reward generation methods.

\subsection{Reward Generation for Cooperative Game}
The proposed framework is applied to generate rewards for training a content generator agent within the multiplayer game environment, RaidEnv II.
To enable LLMs in generating insights tailored to a multiplayer game, we provided a brief description of role differentiation, i.e., one of the multiplayer game design principles \cite{rosen2011managing}.
A set of 32 playtested result values (e.g., survive time, moved distances, damage dealt/taken) are identified as available variables for reward design.
These values are obtained from logs generated by heuristic gameplay agents embedded within the game.
In order to facilitate the alignment process, 1,500 rows of game log data were collected from the environment, and 20 rows of playtested results were randomly sampled to be utilized as code evaluation input.
Self-alignment is set to $N_{align}=5$, reflecting the empirical occurrence of meaningful changes made during the process.
 Prompts used in this work are summarized in this repository\footnote{Prompts are available at https://github.com/bic4907/ChatPCG}.

\section{Experiment}
\subsection{Comparisons}

\begin{itemize}

\item \textbf{Winrate Reward} ($R_{WR}$) ensures that the generated content satisfies the designer-desired winrate.
The L1 norm distance to the goal, $l$, is calculated as $l_t = ||W_{g}-W_{t,c}||_
{L1}$, where $W_{g}$ represents target (goal) winrate and $W_{t,c}$ is the current winrate at time $t$.
The reward at $t$ is then calculated as $r_t = l_{t-1} - l_t$.

\item \textbf{LLM Reward} ($R_{LLM}$) is measured by an LLM-generated reward function.
This function computes the reward by using the playtested results at time $t$ as input.

\item \textbf{Hybrid Reward} ($R_{HYB}$) combines the winrate and LLM-generated rewards with the weights for obtaining the winrate ($w_{WR}$) and LLM ($w_{LLM}$), represented as $R_{HYB} = R_{WR} \cdot w_{WR} + R_{LLM} \cdot w_{LLM}$.

\end{itemize}

The winrate reward was designed based on a previous work \cite{jeon2023raidenv}, while the LLM and hybrid rewards were newly proposed in this paper to consider reward design game characteristics.
Additionally, for comparative analysis, the random (RD) and heuristic (HR) agents utilized in a previous work \cite{jeon2023raidenv} are included.

\subsection{Experiment Setting}
DRL models were trained during 20K steps with three different reward functions; the model hyperparameters are based on previous work \cite{jeon2023raidenv}.
The target winrate ($W_t$) was set to 0.7, and comprehensive results were reported as the average value from three runs.
For generating the LLM-based reward, OpenAI’s \texttt{gpt-4-turbo-2024-04-09} model was employed as the backend language model.
The hybrid reward function weights, $w_{WR}$ and $w_{LLM}$, were empirically determined to be 0.97 and 0.03 to align reward values with similar magnitudes.
For evaluation, character configurations were sampled from trained DRL agents.

\subsection{Evaluation Metrics}
Generated content was evaluated on three criteria: controllability, diversity, and team-build score. 
The definitions of controllability and diversity are described in a previous study \cite{jeon2023raidenv}.
Meanwhile, team-build score is a new evaluation criterion to consider the multiplayer game design and measure role differentiation \cite{rosen2011managing}.
The arrow direction indicates superior values.

\textbf{Controllability (Ctr, $\downarrow$)} measures mean error values between the current and target winrates. A lower error indicates that the model has successfully generated the designer-desired content. It is measured with Eq. \ref{eq:Controllability}, where $N$ denotes the number of samples and $W_{i,c}$ denotes the current winrate of the $i$-th content:
\begin{align}
\label{eq:Controllability}
\text{Ctr} = \frac{1}{N} \sum_{i=1}^{N} ||W_{g}-W_{i,c}||_{L1}
\end{align}

\textbf{Diversity (Div, $\uparrow$)} measures the variability of candidates that satisfy the generation condition.
Valid samples, where the winrate error is lower than the threshold ($\text{abs}(||W_{g}-W_{i,c}||_{L1}) \leq 0.4$), are selected for diversity measurement.
Principal component analysis is applied to the generated character properties to reduce their dimensions from seven to one.
Div calculates the standard deviation of the dimension-reduced values.

\textbf{Team Build Score (Tbs, $\uparrow$)} measures the degree of role differentiation within a team.
It is measured with the following equation, where $T$ denotes the set of agents in a team and $K$ denotes the set of character properties.
$S(a_i, k)$ represents the normalized value of the $i$-th agent's $k$ property in the state $S$.
The mean property value is calculated pairwise among agents.
It is assumed that a higher mean of distances represents distinct character roles in a cooperative setting.
The mean Tbs value for valid samples is calculated with Eq. \ref{eq:tbs}.
\begin{align}
\label{eq:tbs}
\text{Tbs($T$)} = \frac{1}{\binom{|T|}{2}} \sum_{(a_i, a_j) \in T, i < j} \left( \frac{1}{K} \sum_{k \in K} \left| S(a_i,k) - S(a_j,k) \right| \right)
\end{align}

\subsection{Experimental Result}

\begin{table}
\centering
\caption{Comprehensive result for generative models}
\label{tab:evaluation}

\begin{tabular}{@{}llll|lll@{}}
\toprule
\multicolumn{1}{l}{} & PE & Reward & \multicolumn{1}{c}{Ctr ($\downarrow$) ($\pm{}\text{SD}$)} & \multicolumn{1}{c}{Div ($\uparrow$)} & \multicolumn{1}{c}{Tbs ($\uparrow$)} \\ \midrule
RD & - & - & \multicolumn{1}{c}{0.44 ($\pm{}$0.23)} & \multicolumn{1}{c}{0.154} & \multicolumn{1}{c}{0.356} \\
HR & - & - & \multicolumn{1}{c}{0.26 ($\pm{}$0.24)} & \multicolumn{1}{c}{0.053} & \multicolumn{1}{c}{0.103} \\
DRL & - & $R_{WR}$ & \multicolumn{1}{c}{0.22 ($\pm{}$0.05)} & \multicolumn{1}{c}{0.102} & \multicolumn{1}{c}{0.288} \\ \midrule
 & IO & $R_{LLM}$ & \multicolumn{1}{c}{0.27 ($\pm{}$0.04)} & \multicolumn{1}{c}{\textbf{0.346}} & \multicolumn{1}{c}{0.203} \\
DRL & IO & $R_{HYB}$  & \multicolumn{1}{c}{\textbf{0.19 ($\pm{}$0.11)}} & \multicolumn{1}{c}{0.242} & \multicolumn{1}{c}{0.273} \\
(Ours) & CoT & $R_{LLM}$ & \multicolumn{1}{c}{0.39 ($\pm{}$0.23)} & \multicolumn{1}{c}{{0.207}} & \multicolumn{1}{c}{\textbf{0.404}} \\
& CoT & $R_{HYB}$ & \multicolumn{1}{c}{0.21 ($\pm{}$0.13)} & \multicolumn{1}{c}{0.153} & \multicolumn{1}{c}{0.231} \\ \bottomrule
\end{tabular}%

\end{table}

Table \ref{tab:evaluation} presents the descriptive results of various prediction error methods along with different reward functions.
The findings demonstrate that rewards generated by LLMs notably enhance the performance of DRL models, significantly evident in the Div and Tbs score improvements.
Moreover, in terms of Ctr, there is a marginal increment in performance compared to the winrate-only reward.
Contrary to expectations, however, the hybrid rewards did not exhibit benefits of winrate and LLM rewards concurrently.
Concerning prediction error methods, CoT-based $R_{LLM}$ outperforms IO-based $R_{LLM}$ in the Tbs score, achieving the best score across all methods.
The self-alignment process contributes to an improvement in the Tbs score by adapting the reward function to the game.

\subsection{Discussion and Limitation}
The result reveals that LLM-generated reward functions train DRL agents to demonstrate remarkable improvement in generating content tailored to a game characteristic.
However, a notable limitation remains when LLM-generated rewards are incorporated to human-designed reward ($W_{WR}$).
Interestingly, the hybrid reward functions, $R_{HYB}$, outperform the winrate reward function $W_{WR}$ in Ctr; however, the Tbs score decreases by a small value.
While $W_{LLM}$ may have contributed beneficially to the exploration, it is posited that the value of $R_{WR}$ is found to be superior.
Further analysis indicates that heuristically mixed rewards necessitate finely-tuned weights to balance reward signals from several functions, with the LLM-generated reward signal proving superior to the winrate alone.
Consequently, the future work will focus on developing advanced reward generation methods that can satisfy multiple generative objectives through self-alignment.

% \section{Discussion}
% \input{discussion/discussion.tex}

\section{Conclusion and Future Work}
This study presents a LLM-driven automated reward design framework for games.
ChatPCG demonstrates the prowess of LLMs in content-generation literature by designing rewards tailored to a multiplayer game.
The results reveal that LLM-generated rewards effectively improve content quality, further emphasizing the specific aspects of game design.
This approach aims to enhance code quality while enabling multi-objective generation, advancing the capabilities of automated reward design frameworks in game development.
Future research endeavors will delve into a comprehensive analysis of LLM utilizations while also exploring the state-of-the-art PEs.

\section*{Acknowledgments}
This research was supported by Culture, Sports and Tourism R\&D Program through the Korea Creative Content Agency grant funded by the Ministry of Culture, Sports and Tourism in 2022 (Project Name: Development of artificial intelligence-based game simulation technology to support online game content production, Project Number: R2022020070). The asterisk (*) denotes corresponding author.

% This work was also supported by Institute of Information \& communications Technology Planning \& Evaluation (IITP) grant funded by the Korea government (MSIT) (No.2019-0-01842, Artificial Intelligence Graduate School Program (GIST)).

\bibliographystyle{IEEEtran}
\bibliography{references}

% \section{Appendix}
% \input{appendix/appendix.tex}

% \bibitem{b1} G. Eason, B. Noble, and I. N. Sneddon, ``On certain integrals of Lipschitz-Hankel type involving products of Bessel functions,'' Phil. Trans. Roy. Soc. London, vol. A247, pp. 529--551, April 1955.
% \bibitem{b2} J. Clerk Maxwell, A Treatise on Electricity and Magnetism, 3rd ed., vol. 2. Oxford: Clarendon, 1892, pp.68--73.
% \bibitem{b3} I. S. Jacobs and C. P. Bean, ``Fine particles, thin films and exchange anisotropy,'' in Magnetism, vol. III, G. T. Rado and H. Suhl, Eds. New York: Academic, 1963, pp. 271--350.
% \bibitem{b4} K. Elissa, ``Title of paper if known,'' unpublished.
% \bibitem{b5} R. Nicole, ``Title of paper with only first word capitalized,'' J. Name Stand. Abbrev., in press.
% \bibitem{b6} Y. Yorozu, M. Hirano, K. Oka, and Y. Tagawa, ``Electron spectroscopy studies on magneto-optical media and plastic substrate interface,'' IEEE Transl. J. Magn. Japan, vol. 2, pp. 740--741, August 1987 [Digests 9th Annual Conf. Magnetics Japan, p. 301, 1982].
% \bibitem{b7} M. Young, The Technical Writer's Handbook. Mill Valley, CA: University Science, 1989.
% \end{thebibliography}
% \vspace{12pt}

\end{document}